\documentclass[pmlr,twocolumn,10pt]{jmlr} %

\usepackage{booktabs}
\usepackage{siunitx}

\usepackage[switch]{lineno}

\usepackage{multirow}

\usepackage{microtype}
\usepackage{placeins}
\usepackage{float}
\usepackage{stfloats}
\usepackage{enumitem}

\usepackage{soul}
\usepackage{rotating}

\theorembodyfont{\upshape}
\theoremheaderfont{\scshape}
\theorempostheader{:}
\theoremsep{\newline}

\jmlrvolume{259}
\jmlryear{2024}
\jmlrsubmitted{LEAVE UNSET}
\jmlrpublished{LEAVE UNSET}
\jmlrworkshop{Machine Learning for Health (ML4H) 2024} %

 \title[Generalized Prompt Tuning]{Generalized Prompt Tuning: Adapting Frozen Univariate Time Series Foundation Models for Multivariate Healthcare Time Series}

\author{%
\Name{Mingzhu Liu}
\Email{mingzhul@andrew.cmu.edu}\\
\Name{Angela H. Chen}
\Email{angelachen@cmu.edu }\\
\Name{George H.~Chen} \Email{georgechen@cmu.edu}\\
\addr Carnegie Mellon University, Pittsburgh, PA, USA
}

\begin{document}

\maketitle

\begin{abstract}
Time series foundation models are pre-trained on large datasets and are able to achieve state-of-the-art performance in diverse tasks. However, to date, there has been limited work demonstrating how well these models perform in medical applications, where labeled data can be scarce. Further, we observe that currently, the majority of time series foundation models either are univariate in nature, or assume channel independence, meaning that they handle multivariate time series but do not model how the different variables relate. In this paper, we propose a prompt-tuning-inspired fine-tuning technique, {Generalized Prompt Tuning (Gen-P-Tuning)}, that enables us to adapt an existing univariate time series foundation model (treated as frozen) to handle multivariate time series prediction. Our approach provides a way to combine information across channels (variables) of multivariate time series. We demonstrate the effectiveness of our fine-tuning approach against various baselines on two MIMIC classification tasks, and on influenza-like illness forecasting. 
\end{abstract}

\paragraph*{Data and Code Availability}
We use the MIMIC-III dataset \citep{Johnson2016MIMICIIIAF} and an influenza-like illness dataset \citep{wu2021autoformer, cdc}, both of which are publicly available. Our code is available at: \url{https://github.com/Ilovecodingforever/Gen-P-Tuning}

\paragraph*{Institutional Review Board (IRB)}
This research does not require IRB approval as we perform secondary analyses of publicly available datasets.

\setlength{\abovedisplayskip}{4pt plus 2pt}
\setlength{\belowdisplayskip}{4pt plus 2pt}
\setlength{\abovedisplayshortskip}{2pt plus 1.5pt}
\setlength{\belowdisplayshortskip}{2pt plus 1.5pt}

\section{Introduction}

With the rapid development and growing commercial success of large language models (LLMs) recently, there has been a surge of interest in developing similar sorts of foundation models for time series analysis (e.g., \citealt{das2023decoder,gruver2024large, goswami2024moment}). Much like how LLMs are commonly trained on large chunks of the internet spanning many application domains, a number of existing time series foundation models are also trained on a variety of time series data, with the idea that time series across disciplines likely share similar patterns. However, to what extent do such foundation models work for healthcare data such as electronic health records, or public health data such as influenza trends?

For example, electronic health records consist of patient time series, where each time series could be irregularly sampled, have lots of missing entries, and can be of a large number of variables \citep{10.1093/jamia/ocy068}. We suspect that this sort of setting is not well-suited for the current time series foundation models that have been developed. In fact, a major limitation of most time series foundation models that have been developed is that they are \emph{univariate} \citep{ye2024survey}.

In this paper, our main contribution is to show how to adapt existing \emph{unviariate} time series foundation models to handle \emph{multivariate} time series prediction, specifically for both classification and forecasting. We treat the univariate time series foundation model as frozen, and how we adapt it to handle multivariate time series prediction is as a form of \emph{parameter-efficient fine-tuning} (PEFT). PEFT methods have become popular recently for adapting LLMs to handling various datasets (e.g., \citealt{hu2021loralowrankadaptationlarge}). Our proposed PEFT method is a generalization of an existing PEFT method called \emph{prompt tuning}, of which we specifically generalize the variant called \emph{P-tuning v2} by \citet{liu2022ptuningv2prompttuning}. As such, we call our proposed PEFT method \emph{{Generalized Prompt Tuning (Gen-P-Tuning)}}. To be clear, the ``prompt'' here is not a text prompt that a user types. Rather, the prompt for time series forecasting roughly refers to some prefix that we attach to time series (as if we are including some fictitious extra time steps).

We show that {Gen-P-Tuning} is competitive in practice against various fine-tuning baselines in experiments on MIMIC in-hospital mortality prediction and phenotyping, and on influenza-like illness forecasting. %
To the best of our knowledge, our paper is also the first to benchmark various fine-tuning strategies of different time series foundation models on medical datasets. Specifically for our MIMIC classification experiments, we also benchmark against an existing approach called STraTS by \citet{10.1145/3516367}, which is not a fine-tuning method (all our other baselines are fine-tuning methods).

\section{Background}

We first set up some basic notation in Section~\ref{sec:basic-notation}. We then provide an overview of \emph{univariate} time series foundation models in Section~\ref{sec:univariate-fm}. We present this overview at a level of detail that is sufficient for understanding our proposed fine-tuning approach, and that is general enough to encompass a number of existing univariate time series foundation models. We then describe ways in which researchers have adapted univariate time series foundation models for \emph{multivariate} time series prediction in Section~\ref{sec:univariate-to-multivariate}. Lastly, we discuss existing work on developing time series foundation models for medical data in Section~\ref{sec:related-work}. 

\subsection{Basic Notation}
\label{sec:basic-notation}

Throughout our paper, it suffices for us to describe how different time series (foundation) models process a single input time series at a time. In practice, for training and for prediction, the models we describe could trivially process a mini-batch of input time series at once rather than a single time series at a time (for training, standard mini-batch gradient descent could be used for learning trainable parameters).

To this end, we denote a single \emph{multivariate} input time series as $\mathbf{X}\in\mathbb{R}^{C\times T}$, where $C$ is the number of channels/variables, and $T$ is the number of time steps. The multivariate case corresponds to $C\ge2$, whereas in the univariate case ($C=1$), we represent the time series as a 1D array of length $T$ rather than a 2D array of shape $C$-by-$T$.
Importantly, for any dataset of interest, we assume that the number of channels $C$ is fixed.
Across different datasets, it is possible for $C$ to vary. Meanwhile, the number of time steps $T$ can vary across time series even within the same dataset.

We denote the ground truth prediction target as $\mathbf{Y}$. In this paper, we specifically consider two prediction tasks. First, if the prediction task is classification with $M\ge2$ classes, then $\mathbf{Y}\in[0,1]^M$ (the $i$-th entry of $\mathbf{Y}$ indicates the probability of the $i$-th class). Second, if the prediction task is forecasting, then $\mathbf{Y}\in\mathbb{R}^{C \times H}$, where $H$ is the time horizon (i.e., the number of future time steps of the input time series that we aim to predict); in the univariate case, we could of course just represent $\mathbf{Y}$ as a 1D array of length $H$.

\subsection{Univariate Time Series Foundation Models}
\label{sec:univariate-fm}

We now state the general form of the univariate foundation models that our proposed fine-tuning strategy can adapt into multivariate time series predictors. Special cases include, for instance, MOMENT \citep{goswami2024moment}, GPT4TS \citep{zhou2023onefitsall}, and TimesFM \citep{das2023decoder}.

\begin{figure*}[t]
\centering
\includegraphics[width=\linewidth]{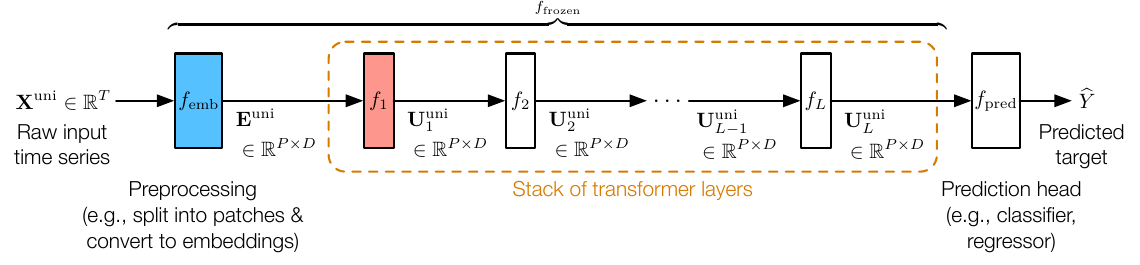}\vspace{-2em}
\caption{Univariate time series foundation model. The predicted target's dimensions depend on the prediction task ($\widehat{Y}\in[0,1]^M$ for $M$-way classification, and $\widehat{Y}\in\mathbb{R}^H$ for forecasting the next $H$ time steps). Note that the preprocessing function $f_{\text{emb}}$ (in blue) and the first transformer layer $f_1$ (in pink) are color-coded for ease of exposition as they reappear later on in Figure~\ref{fig:prompt1}.\vspace{-1em}}
\label{fig:fm}
\end{figure*}

For an input \emph{univariate} time series $\mathbf{X}^{\text{uni}}\in\mathbb{R}^T$ where $T$ is the number of time steps (since this pre-trained foundation model is univariate, there is only a single channel), we assume that the foundation model first applies a preprocessing function $f_{\text{emb}}$ to $\mathbf{X}^{\text{uni}}$ to produce a preprocessed array
\begin{equation}
\mathbf{E}^{\text{uni}}=f_{\text{emb}}(\mathbf{X}^{\text{uni}})\in\mathbb{R}^{P\times D},
\end{equation}
where $P$ is the number of patches (commonly, time series foundation models convert the input time series into patches, so that a raw input time series with~$T$ time steps gets treated instead as a sequence of~$P$ patches \citep{nie2022time})
and $D$ is the embedding dimension. We leave precisely what preprocessing steps are included in the function $f_{\text{emb}}$ unspecified; in practice, this would just depend on whichever univariate foundation model is used.\footnote{For example, prior to chunking the time series into patches, an additional preprocessing step that could be done is reversible instance normalization %
\citep{kim2021reversible}, which standardizes a time series by its mean and variance. An alternative altogether is to preprocess the time series to appear as text data, so that subsequent steps amount to using a large language model \citep{gruver2024large}.}

After preprocessing the raw univariate time series data $\mathbf{X}^{\text{uni}}\in\mathbb{R}^T$ using $f_{\text{emb}}$ to produce $\mathbf{E}^{\text{uni}}\in\mathbb{R}^{P\times D}$, the backbone of the univariate foundation model then applies a sequence of transformer layers one after another. The output from one layer becomes the input to the next layer, for layers 1 to $L$:
\begin{align*}
\mathbf{U}_1^{\text{uni}} &\triangleq f_1(\mathbf{E}^{\text{uni}})\in\mathbb{R}^{P\times D}, \\
\mathbf{U}_2^{\text{uni}} &\triangleq f_2(\mathbf{U}_1^{\text{uni}})\in\mathbb{R}^{P\times D}, \\
&\vdots \\
\mathbf{U}_L^{\text{uni}} &\triangleq f_L(\mathbf{U}_{L-1}^{\text{uni}})\in\mathbb{R}^{P\times D},
\end{align*}
where $f_\ell$ corresponds to the $\ell$-th transformer layer, for $\ell=1,\dots,L$.

Finally, a prediction layer $f_{\text{pred}}$ (sometimes also called a prediction ``head'') is applied to the last transformer layer's output $\mathbf{U}_L^{\text{uni}}$ to produce the final prediction:
\[
\widehat{\mathbf{Y}} = f_{\text{pred}}(\mathbf{U}_L^{\text{uni}}),
\]
where the dimensions of $\widehat{\mathbf{Y}}$ depend on whether we are considering classification or forecasting as the prediction task, as described in Section~\ref{sec:basic-notation}.

In summary, the full univariate foundation model could be represented as the function
\begin{equation}
f \triangleq f_{\text{pred}}\circ \underbrace{(f_L \circ f_{L-1} \circ \cdots \circ f_1 \circ f_{\text{emb}})}_{\triangleq f_{\text{frozen}}},
\label{eq:frozen-plus-head}
\end{equation}
so that $\widehat{\mathbf{Y}} = f(\mathbf{X}^{\text{uni}})=f_{\text{pred}}\big(f_{\text{frozen}}(\mathbf{X}^{\text{uni}})\big)$. This entire process of how raw input time series $\mathbf{X}^{\text{uni}}\in\mathbb{R}^T$ is turned into the final predicted output $\widehat{\mathbf{Y}}$ by the univariate foundation model $f$ is depicted in Figure~\ref{fig:fm}.

Our notation in equation~\eqref{eq:frozen-plus-head} emphasizes that we crucially view all the layers prior to the prediction head as frozen---we will not update parameters of $f_{\text{frozen}}$. The reason we exclude the prediction head $f_{\text{pred}}$ from $f_{\text{frozen}}$ is that when fine-tuning to other datasets, we typically discard the original prediction head $f_{\text{pred}}$ of the foundation model and instead use a new prediction head with trainable parameters.

\paragraph{Generalizing beyond transformers}
Our assumption of the univariate foundation model using repeated transformer layers is mainly because the univariate foundation models we use later in our experiments are transformer-based. However, as it will be apparent when we describe our proposed fine-tuning strategy, the repeated transformer layers could be replaced by some other architecture altogether (i.e., the use of transformers is not actually necessary). This is similar in spirit to how the now-standard Low-Rank Adaptation (LoRA) fine-tuning method \citep{hu2021loralowrankadaptationlarge} is also commonly applied to transformer-based architectures although the key idea of LoRA does not require the use of transformers. %

\subsection{Handling Multivariate Time Series With Channel Independence}
\label{sec:univariate-to-multivariate}

A trivial way of adapting a univariate foundation model for multivariate time series prediction is to assume independence across channels (e.g., \citealt{nie2022time}). Given an input multivariate time series $\mathbf{X}\in\mathbb{R}^{C\times T}$, we separate it out into its different channels' time series $\mathbf{X}_{(1)},\mathbf{X}_{(2)},\dots,\mathbf{X}_{(C)}\in\mathbb{R}^T$. Then we can obtain the final transformer layer's output across the different channels:
\begin{equation}
\mathbf{U}_L^{(c)} \triangleq f_{\text{frozen}}(\mathbf{X}_{(c)})
\qquad\text{for }c=1,\dots,C.
\end{equation}
Finally, assuming that we have training data for the multivariate time series prediction task, we train a prediction head of our choosing that takes in $\mathbf{U}_L^{(1)},\dots,\mathbf{U}_L^{(C)}$ as the inputs (or the average of these across channels as a single input) and outputs the predicted target. As an example, to forecast the next $H$ time steps, we could set the prediction head to be a multilayer perceptron that takes in $\frac{1}{C}\sum_{c=1}^C \mathbf{U}_L^{(c)}$ as input, and outputs a total of $C\cdot H$ numbers (reshaped to be $C$-by-$H$); training can proceed via standard minibatch gradient descent with MSE loss.

\begin{figure*}[t]
\centering
\includegraphics[width=\linewidth]{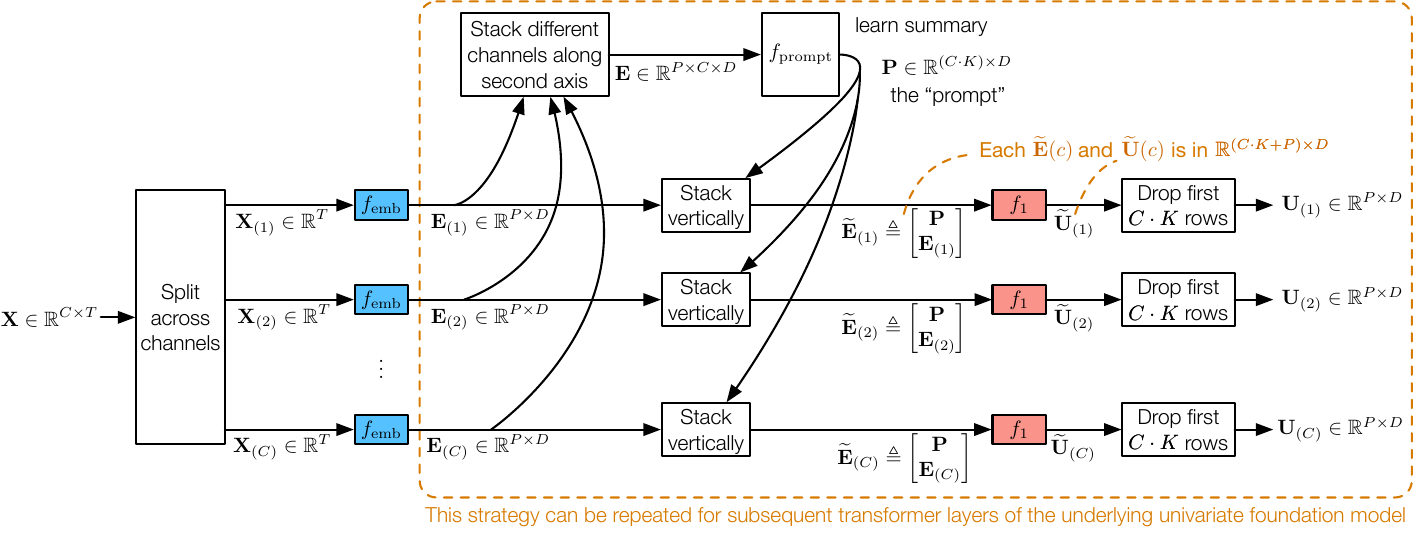}
\vspace{-2em}
\caption{Overview of how the prompt is added to the backbone. Note that there is no subscript in the prompt. The weights are shared across different transformer layers.}
\label{fig:prompt1}
\vspace{-1.5em}
\end{figure*}

\subsection{Time Series Foundation Models for Medical Data}
\label{sec:related-work}

There has not been extensive work on applications of time series foundation models on medical data, especially longitudinal patient data. Time series foundation models are relatively new (compared to LLMs), and the majority of deep learning methods for patient data have been focusing on non-pretrained models such as LSTMs and GRUs \citep{10.1007/978-3-319-31750-2_3, baytas2017patient, che2017rnn, baytas2017patient}.

There is a line of work that aims to build transformer-based foundation models that are pre-trained on EHR data and are able to perform tasks using EHR data. Commonly, the data is encoded as a sequence of tokens, and are used as inputs to a language model. The tokens typically contain a tuple in the form of (timestamp, features) \citep{Yang2023TransformEHRTE, antikainen2023transformers, mcdermott2023event}. For example, CEHR-BERT \citep{pang2021cehrbertincorporatingtemporalinformation} encodes the event at each timestamp as (time, age, clinical concept name), passes them through embedding layers, and feeds it to a BERT (pre-trained language) model. 
Some models are pre-trained on a large cohort of patient data \citep{Yang2023TransformEHRTE, KRALJEVIC2024e281}. However, in order to protect patient privacy, they typically do not release the pre-trained models or data, as opposed to time series foundation models which are typically open-source.

\section{Method}

We now explain how our Gen-P-Tuning method works for adapting a pre-trained univariate time series foundation model for multivariate time series prediction. We treat the univariate foundation model as frozen, and our fine-tuning approach introduces trainable elements. We provide an overview first in Section~\ref{sec:overview}. In this overview, a key component is called a \emph{Prompt Module}, which contains all the trainable elements aside from what is in the final prediction head. We explain the Prompt Module in more detail in Section~\ref{sec:prompt-module}.

\subsection{Overview}
\label{sec:overview}

We give an overview of Gen-P-Tuning first, with an accompanying diagram explaining the high-level steps in Figure~\ref{fig:prompt1}. For ease of exposition, it suffices to explain what happens when the univariate foundation model only uses a single transformer layer ($f_1$ that is shaded in pink in Figure~\ref{fig:fm}). Gen-P-Tuning processes a single multivariate input time series $\mathbf{X}\in\mathbb{R}^{C\times T}$ as follows:
\begin{enumerate}[itemsep=0pt]

\item We separate input $\mathbf{X}\in\mathbb{R}^{C\times T}$ into its different channels' time series $\mathbf{X}_{(1)},\mathbf{X}_{(2)},\dots,\mathbf{X}_{(C)}\in\mathbb{R}^T$. This step is shown in the leftmost part of Figure~\ref{fig:prompt1}.

\item For each channel $c=1,\dots,C$, we preprocess $\mathbf{X}_{(c)}\in\mathbb{R}^T$ using the univariate foundation model's preprocessing function $f_{\text{emb}}$ to obtain the 2D arrays $\mathbf{E}_{(1)},\dots,\mathbf{E}_{(C)}\in\mathbb{R}^{P\times D}$. This step is shown using the blue boxes of Figure~\ref{fig:prompt1} (note that each blue box is the same as the blue box from Figure~\ref{fig:fm}).

\item The 2D arrays $\mathbf{E}_{(1)},\dots,\mathbf{E}_{(C)}\in\mathbb{R}^{P\times D}$ are stacked together treating channels as the second axis to produce a 3D array $\mathbf{E}\in\mathbb{R}^{P\times C\times D}$, as shown in the top box labeled ``Stack different channels along second axis'' in Figure~\ref{fig:prompt1}.

\item We now introduce a trainable neural network $f_{\text{prompt}}$ that takes the 3D array $\mathbf{E}\in\mathbb{R}^{P\times C\times D}$ as input, and outputs a summary 2D array $\mathbf{P}\in\mathbb{R}^{(C\cdot K)\times D}$ that we refer to as the ``prompt''; here $K$ is a user-specified hyperparameter that controls the prompt size. Importantly, the prompt $\mathbf{P}$ should encode summary information \emph{across} channels.

We refer to the function $f_{\text{prompt}}$ as the Prompt Module. We provide the architecture we use for the Prompt Module in Section~\ref{sec:prompt-module}.

\item Next, the prompt $\mathbf{P}\in\mathbb{R}^{(C\cdot K)\times D}$ is ``attached'' to each channel $c$'s preprocessed array $\mathbf{E}_{(c)}$ by vertically stacking $\mathbf{P}$ and $\mathbf{E}_{(c)}$ to produce
\[
\widetilde{\mathbf{E}}_{(c)}
\triangleq
\begin{bmatrix}
\mathbf{P} \\
\mathbf{E}_{(c)}
\end{bmatrix}
\in\mathbb{R}^{(C\cdot K + P)\times D}.
\]
Roughly this could be thought of as adding a prefix of $C\cdot K$ fictitious patches prior to the actual $P$ patches of $\mathbf{E}_{(c)}$. This step corresponds to the boxes labeled ``Stack vertically'' in Figure~\ref{fig:prompt1}.

\item For each channel $c=1,\dots,C$, we feed each array $\widetilde{\mathbf{E}}_{(c)}\in\mathbb{R}^{(C\cdot K + P)\times D}$ as input to the univariate foundation model's first transformer layer $f_1$ to produce the output $\widetilde{\mathbf{U}}_{(c)}\in\mathbb{R}^{(C\cdot K + P)\times D}$. This step corresponds to the pink boxes in Figure~\ref{fig:prompt1}.

\item Note that the first $C\cdot K$ rows of $\widetilde{\mathbf{U}}_{(c)}\in\mathbb{R}^{(C\cdot K + P)\times D}$ correspond to the fictitious patches added in step 5. We now remove these rows as to obtain the output $\mathbf{U}{(c)}\in\mathbb{R}^{P\times D}$, as depicted in the boxes labeled ``Drop first $C\cdot K$ rows'' in Figure~\ref{fig:prompt1}.

Note that $\mathbf{U}_{(1)},\dots,\mathbf{U}_{(C)}\in\mathbb{R}^{P\times D}$ could be viewed as the multivariate output produced using the univariate transformer layer $f_1$ with the help of our Gen-P-Tuning strategy.

\end{enumerate}
If the univariate foundation model has more than one transformer layer, then steps 3--7 could be repeated for the subsequent transformer layers. Supposing for the moment that there is only 1 transformer layer, then after obtaining $\mathbf{U}_{(1)},\dots,\mathbf{U}_{(C)}\in\mathbb{R}^{P\times D}$ from step~7, we would simply train a prediction head in the same manner as stated in Section~\ref{sec:univariate-to-multivariate}, when we covered handling multivariate time series prediction with channel independence. Conceptually, each input to the transformer layer $f_1$ is now provided with information across channels since the prompt $\mathbf{P}$ summarizes information across channels.

We point out two special cases of our approach:
\begin{itemize}

\item (Standard prompt tuning) If the Prompt Module $f_{\text{prompt}}$ is defined to not actually depend on $\mathbf{E}$ and instead just output an array of $(C\cdot K)$-by-$D$ numbers that are all treated as trainable parameters, then we recover a popular prompt tuning strategy called P-tuning v2~\citep{liu2022ptuningv2prompttuning}. This is the main reason we refer to our strategy as Generalized Prompt Tuning.

\item (Channel independence) In the degenerate case where the prompt size hyperparameter $K=0$, so that effectively we do not attach any prefix/prompt $\mathbf{P}$ to each channel's preprocessed array $\mathbf{E}_{(c)}$, then we just recover the same idea as the channel independent strategy of Section~\ref{sec:univariate-to-multivariate}. %

\end{itemize}

\subsection{The Prompt Module\texorpdfstring{ $f_{\text{prompt}}$}{}}
\label{sec:prompt-module}

As our overview above indicates, the Prompt Module $f_{\text{prompt}}$'s main goal is to summarize the different preprocessed time series across channels (a total of $P\times C\times D$ numbers) into a single array $\mathbf{P}\in\mathbb{R}^{(C\cdot K)\times D}$ that notably does not depend on the number of patches $P$. In particular, $f_{\text{prompt}}$ needs to accommodate the possibility that different input time series even within the same dataset could have different numbers of patches~$P$ (and across different datasets, the number of channels~$C$ could vary).

There are many ways to define $f_{\text{prompt}}$. 
We specifically define it to do the following steps:

\begin{enumerate}[itemsep=0pt]

\item We apply a transformer module to $\mathbf{E}\in\mathbb{R}^{P\times C\times D}$ by treating the $P$ different patches as if they are different data points (so that each ``data point'' is in $\mathbb{R}^{C\times D}$, where $C$ is treated as the ``time steps'' by the transformer module). The output per patch is in $\mathbb{R}^{C\times D}$, and we stack these outputs into a single array $\mathbf{P'} \in \mathbb{R}^{P \times C \times D}$.

\item  We use a transformer to map $\mathbf{P'}$ from $\mathbb{R}^{P\times C\times D}$ to $\mathbb{R}^{P \times C\times (K \cdot D)}$, followed by a max pooling operation to map it from $\mathbb{R}^{P \times C\times (K \cdot D)}$ to $\mathbb{R}^{C\times (K \cdot D)}$, and finally a reshape operation to map it from $\mathbb{R}^{C\times (K \cdot D)}$ to $\mathbb{R}^{(K \cdot C)\times D}$. Note that there are other ways of aggregating across the $P$ dimension (to make the output of the Prompt Module not depend on the number of patches). We explain how to instead use an RNN or an MLP in Appendix \ref{apd:prompt} (which includes experimental results with these alternative architectures).

\end{enumerate}

\section{Experiments}

Our experiments aim to show how different fine-tuning strategies for adapting univariate time series foundation models to multivariate time series classification and forecasting work in practice on clinical data and also on a public health dataset. To this end, we specifically consider
univariate time series foundation models, MOMENT \citep{goswami2024moment} and GPT4TS \citep{zhou2023onefitsall} that support both classification and forecasting, and both are special cases of the formulation we presented in Section~\ref{sec:univariate-fm}.\footnote{At a high-level, the major difference between these two foundation models is that MOMENT is based on T5 \citep{raffel2020exploring} whereas GPT4TS is based on GPT2 \citep{radford2019language}.} Moreover, we run an experiment to study the effect of increasing the prompt size hyperparameter $K$, and also point out differences in runtime and the number of parameters used by the different fine-tuning methods.

\paragraph{Datasets}
Classification experiments are performed on MIMIC-III \citep{Johnson2016MIMICIIIAF}, which is a publicly available electronic health records dataset collected from patients in the intensive care units of the Beth Israel Deaconess Medical Center from 2001 to 2012. We follow the benchmark proposed by \citet{harutyunyan2019multitask}. Specifically, we focus on two tasks. The first is a binary classification task of predicting in-hospital mortality based on the first 48 hours of an ICU stay (referred to as \mbox{``MIMIC Mortality''} in our tables later). The second is a multi-class, multi-label classification task, where we classify which of 25 acute care conditions occurred in an ICU stay (``MIMIC Phenotyping''). 

For these two MIMIC classification tasks, to simulate a resource-constrained environment, we use only 1000 randomly sampled patients, 60\% of the data for training, 10\% for validation, and 30\% for testing. In the original benchmark, 17 clinical variables are included.
We also include as an additional variable the number of hours since the time of admission to the ICU. Instead of one-hot encoding as in the benchmark, we encode categorical variables as ordinal values. The time series are irregularly sampled, and missing values are imputed using forward filling when possible or using the normal values suggested by the benchmark otherwise. 

Forecasting experiments are performed on an influenza-like illness dataset \citep{wu2021autoformer, cdc}. This is a weekly sampled dataset with 7 channels/variables. The dataset is split into 60\% training, 10\% validation, and 30\% testing. The forecasting horizon is 60 weeks. We forecast all 7 variables in the dataset.\footnote{We point out that \citet{goswami2024moment} also presented forecasting results on this dataset but they report experimental results only on forecasting one of the 7 variables (``OT''), so the numbers they get are not directly comparable to ours.} %

\begin{table*}[t]
\centering
\caption{MIMIC Mortality test set scores (mean $\pm$ std.~dev.~over 5 random seeds). For each univariate foundation model, per column we bold whichever score is highest and underline the second-best score. Note that STraTS is not a univariate foundation model that we fine-tune, so it is provided as a non-foundation-model baseline.\vspace{-.5em}}
\label{tab:mortality}%
{
\small
     \begin{tabular}{llcccc} \toprule
    Model & Fine-Tuning Method & Raw Accuracy & AUROC  & F1  & AUPRC \\ \midrule
    \multirow{4}{*}{MOMENT}  
     &  Full  &  \textbf{0.891} $\pm$ 0.012 & 0.687 $\pm$ 0.020  & 0.508 $\pm$ 0.044  & 0.255 $\pm$ 0.038   \\
     &  LoRA  &  0.875 $\pm$ 0.021  & 0.720 $\pm$ 0.019  & 0.573 $\pm$ 0.050  & 0.272 $\pm$ 0.025   \\
     &  Linear Probing  &  0.878 $\pm$ 0.013  & \underline{0.730} $\pm$ 0.035 & 0.544 $\pm$ 0.043  & 0.260 $\pm$ 0.018  \\
     &  Prompt Tuning  &  \underline{0.883} $\pm$ 0.012 & 0.724 $\pm$ 0.035  & \underline{0.576} $\pm$ 0.058 & \underline{0.274} $\pm$ 0.020  \\
     &  Gen-P-Tuning  &  0.881 $\pm$ 0.005  & \textbf{0.754} $\pm$ 0.021 & \textbf{0.591} $\pm$ 0.031 & \textbf{0.292} $\pm$ 0.026 \\
    \cmidrule(lr){1-6}
    \multirow{4}{*}{GPT4TS}  
     &  Full  &  0.886 $\pm$ 0.019  & \textbf{0.743} $\pm$ 0.018 & 0.524 $\pm$ 0.052  & \textbf{0.309} $\pm$ 0.023 \\
     &  LoRA  &  0.871 $\pm$ 0.017  & 0.708 $\pm$ 0.056  & \textbf{0.588} $\pm$ 0.028 & 0.254 $\pm$ 0.024  \\
     &  Linear Probing  &  0.859 $\pm$ 0.015  & \underline{0.737} $\pm$ 0.033 & \underline{0.584} $\pm$ 0.037 & \underline{0.265} $\pm$ 0.037\\
     &  Prompt Tuning  &  \textbf{0.891} $\pm$ 0.013 & 0.689 $\pm$ 0.062  & 0.471 $\pm$ 0.004  & 0.236 $\pm$ 0.022  \\
     &  Gen-P-Tuning  &  \underline{0.887} $\pm$ 0.016 & 0.708 $\pm$ 0.025  & 0.499 $\pm$ 0.033  & 0.255 $\pm$ 0.038   \\
     \cmidrule(lr){1-6} 
     \multirow{1}{*}{STraTS} && 0.900 $\pm$ 0.000 &	0.601 $\pm$ 0.039	& 0.474 $\pm$ 0.000	& 0.159 $\pm$ 0.039 \\
     \bottomrule
    \end{tabular}}
\vspace{-.75em}
\end{table*}
\begin{table*}[!t]
\caption{MIMIC Phenotyping test set scores (mean $\pm$ std.~dev.~over 5 random seeds). This table uses the same formatting as Table~\ref{tab:mortality} (in terms of what bolding and underlining mean).\vspace{-.75em}}
  \label{tab:phenotyping}%
{
\small
        \begin{tabular}{llcccc} \toprule
        Model & Fine-Tuning Method & Raw Accuracy & AUROC & F1 & AUPRC\\ \midrule
        \multirow{4}{*}{MOMENT} 
         &  Full  &  \underline{0.832} $\pm$ 0.007  & \underline{0.643} $\pm$ 0.019 & 0.070 $\pm$ 0.024  & \underline{0.276} $\pm$ 0.021 \\
         &  LoRA  &  \underline{0.832} $\pm$ 0.007 & 0.640 $\pm$ 0.025  & \underline{0.085} $\pm$ 0.023 & 0.273 $\pm$ 0.027   \\
         &  Linear Probing  &  0.830 $\pm$ 0.006  & 0.631 $\pm$ 0.026  & 0.071 $\pm$ 0.031  & 0.264 $\pm$ 0.022   \\
         &  Prompt Tuning  &  \underline{0.832} $\pm$ 0.008  & 0.634 $\pm$ 0.012  & 0.069 $\pm$ 0.036  & 0.268 $\pm$ 0.015  \\
         &  Gen-P-Tuning  &  \textbf{0.835} $\pm$ 0.004 & \textbf{0.666} $\pm$ 0.015 & \textbf{0.135} $\pm$ 0.017 & \textbf{0.294} $\pm$ 0.012  \\
         \cmidrule(lr){1-6}
        \multirow{4}{*}{GPT4TS} 
         &  Full  &  0.823 $\pm$ 0.009  & 0.593 $\pm$ 0.014  & 0.060 $\pm$ 0.028  & \underline{0.234} $\pm$ 0.014  \\
         &  LoRA  &  0.801 $\pm$ 0.015  & \underline{0.596} $\pm$ 0.023 & \underline{0.107} $\pm$ 0.024 & \textbf{0.241} $\pm$ 0.015  \\
         &  Linear Probing  &  0.789 $\pm$ 0.009  & 0.555 $\pm$ 0.016  & \textbf{0.129} $\pm$ 0.029 & 0.213 $\pm$ 0.015  \\
         &  Prompt Tuning  &  \underline{0.831} $\pm$ 0.010 & 0.581 $\pm$ 0.012  & 0.024 $\pm$ 0.017  & 0.227 $\pm$ 0.014  \\
         &  Gen-P-Tuning  &  \textbf{0.832} $\pm$ 0.003 & \textbf{0.599} $\pm$ 0.010 & 0.020 $\pm$ 0.009  & 0.231 $\pm$ 0.010  \\
          \cmidrule(lr){1-6} 
         \multirow{1}{*}{STraTS} && 0.835 $\pm$ 0.008 &	0.573 $\pm$ 0.020 &	0.000 $\pm$ 0.000 &	0.217 $\pm$ 0.016 \\
         \bottomrule
        \end{tabular}
          }
\vspace{.25em}
\caption{Influenza-like illness forecasting test set scores (mean $\pm$ std.~dev.~over 5 random seeds). For each univariate foundation model, per column we bold whichever score is best and underline the second-best score.\vspace{-.5em}}
\label{tab:forecasting}
\centering
\small
        \begin{tabular}{llcc} 
        \toprule
        Model & Fine-Tuning Method & MSE & MAE\\ \midrule
        \multirow{4}{*}{MOMENT} 
         &  Full  &  3.199 $\pm$ 0.102  & 1.262 $\pm$ 0.022  \\
         &  LoRA  &  3.109 $\pm$ 0.021  & \underline{1.176} $\pm$ 0.006 \\
         &  Linear Probing  &  \textbf{2.622} $\pm$ 0.036 & \textbf{1.159} $\pm$ 0.011 \\
         &  Prompt Tuning  &  \underline{2.918} $\pm$ 0.047 & 1.200 $\pm$ 0.009  \\
         &  Gen-P-Tuning  &  3.083 $\pm$ 0.080  & 1.200 $\pm$ 0.016  \\
         \cmidrule(lr){1-4}
        \multirow{4}{*}{GPT4TS} 
         &  Full  &  3.219 $\pm$ 0.093  & 1.240 $\pm$ 0.012   \\
         &  LoRA  &  3.247 $\pm$ 0.337  & \underline{1.230} $\pm$ 0.089  \\
         &  Linear Probing  &  3.202 $\pm$ 0.303  & 1.232 $\pm$ 0.072  \\
         &  Prompt Tuning  &  \underline{3.105} $\pm$ 0.429 & 1.253 $\pm$ 0.098   \\
         &  Gen-P-Tuning  &  \textbf{2.939} $\pm$ 0.378 & \textbf{1.209} $\pm$ 0.092 \\
         \bottomrule
        \end{tabular}
\vspace{.25em}
\caption{Validation loss of linear probing, prompt tuning, and Gen-P-Tuning on MOMENT mortality prediction task (mean $\pm$ std.~dev.~over 5 random seeds).}\vspace{-.5em}
\label{tab:validation}
  {%
\centering
\small
\begin{tabular}{lc}
  \toprule
Fine-Tuning Method & Validation Loss \\
  \midrule
Linear Probing & 0.396 $\pm$ 0.094	\\
Prompt Tuning &	0.397 $\pm$ 0.100 \\
Gen-P-Tuning	& 0.347 $\pm$ 0.064	 \\
  \bottomrule
\end{tabular}
}
\vspace{-0.75em}
\end{table*}

\paragraph{Baselines}
We compare Gen-P-Tuning to the following fine-tuning baselines:

\begin{enumerate}[itemsep=0pt]
    \item Full fine-tuning: all parameters are updated.
    \item LoRA \citep{hu2021loralowrankadaptationlarge}: this baseline keeps the pre-trained weight matrices for the transformer blocks frozen but allows for each block to be ``updated'' by a low-rank matrix in the following manner. Let $W$ denote the weight matrix for a specific transformer block. Then we simply replace $W$ with $W_{\text{new}} = W + AB$, where matrices  $A$ and $B$ are low rank. LoRA fine-tuning corresponds to leaving the original $W$ fixed and only learning the values in $A$ and $B$.
    \item Linear probing: this baseline is precisely the channel-independent strategy mentioned in Section~\ref{sec:univariate-to-multivariate}. For MOMENT, the only learnable part is now the prediction head. For GPT4TS, this includes the input embedding layer (since it is not pre-trained) and the prediction head. 
    \item P-tuning v2 (referred to simply as ``Prompt Tuning'' in our tables later), where $\mathbf{P}$ is an array of trainable parameters.
\end{enumerate}
As we already stated at the end of Section~\ref{sec:overview}, the channel-independent strategy (linear probing) and standard prompt tuning (P-tuning v2) could be viewed as special cases of our Generalized Prompt Tuning approach. However, in our experiments to follow, for simplicity, the results we show for Gen-P-Tuning are specific to when we define the Prompt Module (Section~\ref{sec:prompt-module}) in a nontrivial manner so that it does not simply reduce to either linear probing or P-tuning v2. In practice, one could of course tune our Gen-P-Tuning strategy (e.g., based on a validation set evaluation metric) to choose between a nontrivial Prompt Module, a trivial Prompt Module that just outputs a trainable array of numbers (resulting in P-tuning v2), or no Prompt Module (resulting in linear probing).

For MIMIC experiments, we also include a baseline that is not a fine-tuning approach. Specifically, we use STraTS \citep{10.1145/3516367}, a transformer-based model developed on MIMIC-III that applies self-supervised pretraining (forecasting the values in the next two hours). It encodes EHR data as a sequence of triplets (timestamp, variable name, variable value). We pretrain the model using 1000 random samples from the benchmark, and then perform the classification tasks using the same samples.

\paragraph{Evaluation metrics}
We use accuracy, area under the receiver operating characteristic curve (AUC), F1, and area under the precision-recall curve (AUPRC) for classification tasks. The macro variant of each metric is used for MIMIC Phenotyping.\footnote{In the classification tasks, we point out that raw accuracy alone is not a representative metric since there is a class imbalance in MIMIC-III tasks (negative to positive ratio is 7:1 for mortality prediction, and 5:1 on average for phenotype classification).} We use mean squared error (MSE) and mean absolute error (MAE) for the forecasting task.

\paragraph{Main benchmark findings}
We present MIMIC mortality classification performance in Table~\ref{tab:mortality}, MIMIC phenotype classification performance in Table~\ref{tab:phenotyping}, and influenza-like illness forecasting performance in Table~\ref{tab:forecasting}.

There is no method that always performs the best. Gen-P-Tuning is often among the best-performing ones. Linear probing sometimes performs very well, which suggests that a channel-independent strategy is sometimes sufficient. Indeed, some existing work has demonstrated that channel-independence sometimes performs well. For example, PatchTST \citep{nie2022time}, which is a channel-independent method, has been shown to outperform ``channel-mixing'' where channels are concatenated before being fed into the model.

Full fine-tuning sometimes does not perform well, especially for forecasting experiments. This might be a sign of catastrophic forgetting \citep{goodfellow2013empirical}, which is commonly observed in low-data regimes with deep networks. There has not been previous work on catastrophic forgetting of time series foundation models, but it has been observed in models such as LSTMs \citep{10.1007/978-3-030-30484-3_56}.

\begin{table*}[t]
\centering
\caption{MOMENT Mortality test set scores (mean $\pm$ std.~dev.~over 5 random seeds) with varying prompt size. As a reminder, larger prompt size hyperparameter $K$ corresponds to a larger prompt size.\vspace{-.5em}}
\label{tab:mimic_prefix}
{\small
\begin{tabular}{ccccccc} \toprule
Prompt size hyperparameter $K$  & Raw Accuracy & AUROC  & F1 & AUPRC \\ \midrule
1 &   0.879 $\pm$ 0.011   &  0.759 $\pm$ 0.019 &  0.574 $\pm$ 0.015   &  0.293 $\pm$ 0.037    \\
2 &  0.871 $\pm$ 0.007   &  0.748 $\pm$ 0.033 &   0.593 $\pm$ 0.036   &  0.297 $\pm$ 0.056     \\
4 & 0.881 $\pm$ 0.005  &  {0.754} $\pm$ 0.021 &  {0.591} $\pm$ 0.031 & {0.292} $\pm$ 0.026  \\ \bottomrule
\end{tabular}}
\caption{Number of trainable parameters in mortality prediction experiments.}\vspace{.25em}
\label{tab:param}
{\small\setlength{\tabcolsep}{3.25pt}
\begin{tabular}{lccccccccc}
  \toprule
Model & Full fine-tune & LoRA & Linear Probing & Prompt Tuning & Gen-P-Tuning \\
  \midrule
MOMENT & 341,651,993 (100\%) & 666,649 (0.2\%) & 411,673 (0.1\%) & 685,465 (0.2\%) & 619,777  (0.2\%)  \\
GPT4TS & 60,985,345 (100\%) & 1,132,033 (1.9\%) & 139,777 (0.2\%) & 1,270,881 (2.1\%) & 1,646,593 (2.7\%)\\ \bottomrule
\end{tabular}\vspace{-.5em}}
\vspace{.5em}
\caption{Runtime in seconds of all fine-tuning strategies (one experimental repeat).}%
\label{tab:time}\vspace{.25em}%
{\small%
\begin{tabular}{lcccccccc}
  \toprule
\multirow{2}{*}{Model} & \multirow{2}{*}{Experiment}  & Full & \multirow{2}{*}{LoRA}  & Linear & Prompt & \multirow{2}{*}{Gen-P-Tuning} \\
&& fine-tune && Probing & Tuning & \\
  \midrule
\multirow{3}{*}{MOMENT} & MIMIC Mortality &	1218	& 1025	& 418 & 706	& 3317 \\
& MIMIC Phenotyping	& 1287	& 1084	& 421 & 697	& 3507 \\
& Forecasting	& 46	& 42 &	32	& 38	& 70 \\
  \midrule
\multirow{3}{*}{GPT4TS} & MIMIC Mortality &	130 &	119	& 107 & 430	& 913 \\
& MIMIC Phenotyping	& 137 &	124 &	112 & 432	& 943 \\
& Forecasting	& 7 &	7 &	6 &	29 &	43 \\ \bottomrule
\end{tabular}}
\vspace{-0.75em}
\end{table*}

As a reminder, linear probing and standard prompt tuning can be thought of as special cases of Generalized Prompt Tuning, where we set hyperparameters differently. We can choose between these special cases by whichever one achieves the best validation loss. As an illustration of this, for MIMIC mortality prediction, using the MOMENT foundation model, we report the validation losses of linear probing, standard prompt tuning, and Gen-P-Tuning (with a nontrivial Prompt Module) in Table \ref{tab:validation}. In this case, Gen-P-Tuning achieves the lowest validation loss, and on the true test set, it does indeed outperform the simpler special cases.

\paragraph{Prompt size}
We demonstrate the effect of increasing the prompt size on the MIMIC-III mortality prediction task (Table~\ref{tab:mimic_prefix}). For this task, performance generally becomes better as the prompt size increases.

\paragraph{Computational efficiency}
To give a sense of the number of trainable parameters for the different fine-tuning methods, we report these numbers specifically for the MIMIC-III mortality prediction in \tableref{tab:param}.

\paragraph{Runtime}
We present the runtime in seconds in \tableref{tab:time}. The main reason why both prompt tuning and Gen-P-Tuning can take more time than full fine-tuning, LoRA, and linear probing is that the Prompt Module needs to be trained and it depends on all channels. Furthermore, the backbones of MOMENT and GPT4TS are implemented in Huggingface as is the LoRA fine-tuning method. It is possible that Huggingface's implementation of these are optimized for faster training, whereas we did not prioritize computational efficiency in our implementations of Prompt Tuning and Gen-P-Tuning.

\section{Discussion}

In this paper, we demonstrate the applicability of fine-tuning methods to time series foundation models on disease surveillance and electronic health records data. Moreover, we propose a prompt-tuning method that fine-tunes univariate time series foundation models for multivariate time series classification and forecasting by adding a Prompt Module that combines information across channels. %

We highlight some limitations of our work that in turn point toward directions of future research.
First, we only evaluated the performance of models on two datasets, with only a fairly limited collection of feature types. The influenza-like illness dataset already resembles many other time series forecasting tasks so that we would expect time series foundation models to work well for it. As for MIMIC, we reused the setup by \citet{harutyunyan2019multitask}, which only considers a relatively small set of features. There are other modalities common in healthcare, such as waveforms, which are recorded more frequently and thus have a longer context available. Importantly, healthcare time series also routinely consist of categorical variables that change over time. We suspect that time-varying categorical variables do not closely resemble the sort of time series that time series foundation models are typically trained on. However, we have not investigated how fine-tuning time series foundation models copes with these categorical variables.

A separate direction that we have not explored is interpreting how the learned Prompt Module combines information across channels. While attention weights of the Prompt Module could be visualized, there is debate on whether attention weights are interpretable (e.g., \citealt{serrano2019attention}). A direction that could be promising is to synthetically combine univariate time series to form multivariate time series (so we know how channels are combined). We can then check whether the Prompt Module can recover the ground truth channel mixing strategy.

\acks{We thank the anonymous reviewers for their helpful feedback. G.~H.~Chen is supported by NSF CAREER award \#2047981.}

\bibliography{references}

\appendix

\section{Implementation Details}\label{apd:implementation}

Experiments were run with NVIDIA RTX A6000, Python 3.11.5, Pytorch 2.4.0, Huggingface-hub 0.24.0, and MOMENT-1-large.

All experiments were run with the following hyperparameters:

\begin{enumerate}[itemsep=0pt]
    \item Number of epochs: 10
    \item Scheduler: OneCycleLR
    \item Optimizer: AdamW
    \item Learning rate: $5\times10^{-5}$
    \item Max learning rate: 0.01
    \item Weight decay: 0.05
    \item Loss function: MSE for forecasting, binary cross-entropy for classification
    \item Prompt size $K$: 16 for forecasting dataset, 4 for classification
    \item LoRA hyperparameters:
    \begin{itemize}[itemsep=0pt]
        \item Attention dimension: 1 (attention dimension was chosen to make the number of trainable parameters of LoRA close to that of Gen-P-Tuning.)
        \item Alpha: 16
        \item Dropout: 0.1
    \end{itemize}
\end{enumerate}

Other parameters are taken from the source code or the paper of MOMENT. We did not perform any hyperparameter tuning.

\begin{table*}[t]
\centering
\caption{MIMIC Mortality test set scores (mean $\pm$ std.~dev.~over 5 random seeds). For each univariate foundation model, per column we bold whichever score is highest.
\vspace{-.75em}}
\label{tab:mortality_prompt}%
{
\small
     \begin{tabular}{llcccc} \toprule
    Model & Aggregation Method & Raw Accuracy & AUROC  & F1  & AUPRC \\ \midrule
    \multirow{3}{*}{MOMENT}  
    &  MLP  & \textbf{0.882} $\pm$ 0.010 & 0.750 $\pm$ 0.025 & 0.553 $\pm$ 0.044 & {0.280} $\pm$ 0.017 \\
    & RNN & 0.881 $\pm$ 0.010   &  \textbf{0.755} $\pm$ 0.027   &  0.584 $\pm$ 0.029   &  \textbf{0.304} $\pm$ 0.029 \\
     &  Transformer  &  0.881 $\pm$ 0.005  & {0.754} $\pm$ 0.021 & \textbf{0.591} $\pm$ 0.031 & {0.292} $\pm$ 0.026 \\
    \cmidrule(lr){1-6}
    \multirow{3}{*}{GPT4TS}  
    &  MLP  & \textbf{0.892} $\pm$ 0.014 & 0.685 $\pm$ 0.066 & 0.479 $\pm$ 0.017 & 0.226 $\pm$ 0.051 \\
    & RNN  &  0.890 $\pm$ 0.011   &  \textbf{0.716} $\pm$ 0.030   &  0.489 $\pm$ 0.032   &  0.253 $\pm$ 0.026  \\
     &  Transformer  &  {0.887} $\pm$ 0.016 & 0.708 $\pm$ 0.025  & \textbf{0.499} $\pm$ 0.033  & \textbf{0.255} $\pm$ 0.038   \\
     \bottomrule
    \end{tabular}}
\caption{MIMIC Phenotyping test set scores (mean $\pm$ std.~dev.~over 5 random seeds). For each univariate foundation model, per column we bold whichever score is highest.
 \vspace{-.75em}}
  \label{tab:phenotyping_prompt}%
{
\small
        \begin{tabular}{llcccc} \toprule
        Model & Aggregation Method & Raw Accuracy & AUROC & F1 & AUPRC\\ \midrule
        \multirow{3}{*}{MOMENT} 
        &  MLP  & 0.830 $\pm$ 0.006 & 0.640 $\pm$ 0.018 & {0.079} $\pm$ 0.030 & 0.270 $\pm$ 0.015 \\
        & RNN  &  \textbf{0.837} $\pm$ 0.004   &  \textbf{0.669} $\pm$ 0.013   &  0.129 $\pm$ 0.024   &  \textbf{0.297} $\pm$ 0.013  \\
         &  Transformer &  {0.835} $\pm$ 0.004 & {0.666} $\pm$ 0.015 & \textbf{0.135} $\pm$ 0.017 & {0.294} $\pm$ 0.012  \\
         \cmidrule(lr){1-6}
        \multirow{3}{*}{GPT4TS} 
        & MLP & \textbf{0.832} $\pm$ 0.005 & 0.581 $\pm$ 0.018 & 0.026 $\pm$ 0.024 & 0.230 $\pm$ 0.016  \\
        & RNN  &  0.831 $\pm$ 0.006   &  0.592 $\pm$ 0.008   &  \textbf{0.037} $\pm$ 0.013   &  0.229 $\pm$ 0.007  \\
         &  Transformer  &  \textbf{0.832} $\pm$ 0.003 & \textbf{0.599} $\pm$ 0.010 & 0.020 $\pm$ 0.009  & \textbf{0.231} $\pm$ 0.010  \\
         \bottomrule
        \end{tabular}
          }
\caption{Influenza-like illness forecasting test set scores (mean $\pm$ std.~dev.~over 5 random seeds). For each univariate foundation model, per column we bold whichever score is best.
\vspace{-.75em}}
\label{tab:forecasting_prompt}
\centering
\small
        \begin{tabular}{llcc} 
        \toprule
        Model & Aggregation Method & MSE & MAE\\ \midrule
        \multirow{3}{*}{MOMENT} 
        & MLP & \textbf{2.718} $\pm$ 0.049 & \textbf{1.154} $\pm$ 0.014 \\
        & RNN  &  3.044 $\pm$ 0.081   &  1.210 $\pm$ 0.017  \\
         &  Transformer  &  3.083 $\pm$ 0.080  & 1.200 $\pm$ 0.016  \\
         \cmidrule(lr){1-4}
        \multirow{3}{*}{GPT4TS} 
        & MLP & {3.196} $\pm$ 0.556 & 1.253 $\pm$ 0.104 \\
        & RNN  &  \textbf{2.721} $\pm$ 0.678   &  \textbf{1.140} $\pm$ 0.145 \\
         &  Transformer  &  {2.939} $\pm$ 0.378 & {1.209} $\pm$ 0.092 \\
         \bottomrule
        \end{tabular}
\end{table*}

\section{Aggregating Across Patches in the Prompt Module}\label{apd:prompt}

The input to the Prompt Module has a size that scales with the number of patches whereas the output of the Prompt Module does not depend on the number of patches. As a reminder (Section \ref{sec:prompt-module}'s step 1), the Prompt Module first embeds $\mathbf{E}$ (using an transformer) to obtain the array $\mathbf{P'}\in \mathbb{R}^{P \times C \times D}$. At this point, we want to summarize $\mathbf{P'}$ into an array that does not depend on $P$. There are many ways of doing this summarization that is fundamentally about aggregating information from $\mathbf{P}'$ across patches. Some examples include transformers, recurrent neural networks (RNN), and multi-layer perceptions (MLP). Transformers and RNNs are able to process sequences that vary in length. MLPs could also be applicable since many time series foundation models pad or truncate input sequences so that they are all the same length.

In more detail, we could aggregate across patches using the following three different approaches:

\begin{enumerate}[itemsep=0pt]
    \item Transformer: A transformer strategy has already been presented in Section \ref{sec:prompt-module}'s step 2.
    \item RNN: We use an RNN to map $\mathbf{P'}$ from $\mathbb{R}^{P\times C\times D}$ to $\mathbb{R}^{P \times C\times (K \cdot D)}$. We use the output at the last timestamp, and then a reshape operation to map it from $\mathbb{R}^{C\times (K \cdot D)}$ to $\mathbb{R}^{(K \cdot C)\times D}$. 
    \item MLP: We use a MLP to map $\mathbf{P'}$ from $\mathbb{R}^{P\times C\times D}$ to $\mathbb{R}^{K \times C\times D}$, then followed by a flatten operation to map it from $\mathbb{R}^{K\times C\times D}$ to $\mathbb{R}^{(K \cdot C)\times D}$. 
\end{enumerate}

We present the test performance of MIMIC Mortality (Table \ref{tab:mortality_prompt}), MIMIC Phenotyping (Table \ref{tab:phenotyping_prompt}), and forecasting (Table \ref{tab:forecasting_prompt}) using transformer, RNN, and MLP aggregation strategies.

\end{document}